\documentclass[11pt]{article}

\usepackage{booktabs}       
\usepackage{amsfonts}       
\usepackage{nicefrac}       
\usepackage{microtype}      

\usepackage{fncylab,enumerate,wrapfig,placeins}
\usepackage{bbm}
\usepackage{graphicx}
\usepackage{enumitem}  
\usepackage{amsmath,amsthm,amsfonts,amssymb,xcolor}  
\usepackage{pifont}

\usepackage{titlesec}

\usepackage{scalerel}
\usepackage{stackengine}
\stackMath

\providecommand{\mathlarger}[1]{#1}

\usepackage[margin=1in]{geometry}

\usepackage{amsmath, amssymb, amsthm}

\newtheorem{theorem}{Theorem}[section]
\newtheorem{lemma}[theorem]{Lemma}
\newtheorem{proposition}[theorem]{Proposition}

\usepackage{algorithm}
\usepackage{algpseudocode}

\usepackage{graphics}
\usepackage{subfigure}
\graphicspath{{../figures/}}

\usepackage{enumerate, wrapfig, placeins, verbatim}
\usepackage{optidef}

\usepackage{textgreek, upgreek, bm}

\usepackage[hidelinks]{hyperref}

\usepackage{cleveref}

\Crefname{corollary}{Corollary}{Corollaries}
\Crefname{eqnarray}{eq.}{eqs.}
\Crefname{equation}{eq.}{eqs.}
\Crefname{figure}{Fig.}{Figs.}
\Crefname{wrapfigure}{Fig.}{Figs.}
\Crefname{tabular}{Tab.}{Tabs.}
\Crefname{table}{Tab.}{Tabs.}
\Crefname{lemma}{Lemma}{Lemmas}
\Crefname{theorem}{Thm.}{Thms.}
\Crefname{definition}{Definition}{Definitions}
\Crefname{section}{Section}{Sections}
\Crefname{proposition}{Prop.}{Propositions}

\newlength{\noteWidth}
\long\def\notes#1{\ifinner
	{\tiny #1}
\else
\marginpar{\parbox[t]{\noteWidth}{\raggedright\tiny #1}}
\fi}
\def\notes#1{}

\def\urls#1{{\footnotesize\url{#1}}}

\makeatletter
\newcommand\gobblepars{%
    \@ifnextchar\par%
 {\expandafter\gobblepars\@gobble}%
{}}
\makeatother

\def\wham#1{\smallbreak\pagebreak[3]%
	\noindent\textup{\textbf{#1}}\ \ \gobblepars}

\def\whamrm#1{\smallbreak\pagebreak[3]%
\noindent\textrm{#1}\ \ \gobblepars}

\def\whamb{\wham{$\bullet$}}

\def\mindex#1{\index[main]{#1}}

\DeclareFontFamily{U}{mathx}{\hyphenchar\font45}
\DeclareFontShape{U}{mathx}{m}{n}{<-> mathx10}{}
\DeclareSymbolFont{mathx}{U}{mathx}{m}{n}
\DeclareMathAccent{\widebar}{0}{mathx}{"73}

\DeclareFontFamily{U}{bbold}{}
\DeclareFontShape{U}{bbold}{m}{n}
   {  <-5.5> bbold5 <5.5-6.5> bbold6 <6.5-7.5> bbold7
      <7.5-8.5> bbold8 <8.5-9.5> bbold9 <9.5-11> bbold10
      <11-15> bbold12 <15-> bbold17 }{}
\DeclareSymbolFont{bbold}{U}{bbold}{m}{n}
\DeclareSymbolFontAlphabet{\mathbbold}{bbold}

\makeatletter
\newcommand{\bigblock}[2][\LARGE]{%
  \mkern4mu
  \vcenter{\kern.3ex\hbox{#1$\m@th\mathstrut\mathbbold{#2}$}}%
  \mkern1mu
}
\makeatother

\newcommand{\ind}{{\mathchoice{\bigblock[\large]{1}}%
{\mathlarger{\mathbbold{1}}}
{\mathbbold{1}}%
{\mathbbold{1}}}}

\renewcommand{\ind}{\mathbbold{1}}

\def\Obj{\Upgamma}  

\def\odestate{\upvartheta}

\def\Tdiff{\mathcal{D}}

\def\fee{\upphi}

\def\feex{{\breve{\fee}}} 

\def\elig{\zeta}

\def\uH{\underline{H}}

\def\uQ{\underline{Q}}

\def\uQfee{\underline{Q}_{\raise 1.25pt\hbox{\tiny$\fee$}} }

\def\disc{\gamma}

\newcommand{\bbblot}{\raise1pt\hbox{\vrule height .4ex width .4ex depth .05ex}}

\long\def\defbox#1{\framebox[.9\hsize][c]{\parbox{.85\hsize}{%
\parindent=0pt
\baselineskip=12pt plus .1pt      
\parskip=6pt plus 1.5pt minus 1pt 
 #1}}}

\long\def\beginbox#1\endbox{\subsection*{}%
\hbox{\hspace{.05\hsize}\defbox{\medskip#1\bigskip}}%
\subsection*{}}

\def\endbox{}

 \def\archival#1{} 

\def\FRAC#1#2#3{\genfrac{}{}{}{#1}{#2}{#3}}

\def\ddt{{\mathchoice{\FRAC{1}{d}{dt}}%
{\FRAC{1}{d}{dt}}%
{\FRAC{3}{d}{dt}}%
{\FRAC{3}{d}{dt}}}}

\def\ddtp{{\mathchoice{\FRAC{1}{d^{\hbox to 2pt{\rm\tiny +\hss}}}{dt}}%
{\FRAC{1}{d^{\hbox to 2pt{\rm\tiny +\hss}}}{dt}}%
{\FRAC{3}{d^{\hbox to 2pt{\rm\tiny +\hss}}}{dt}}%
{\FRAC{3}{d^{\hbox to 2pt{\rm\tiny +\hss}}}{dt}}}}

\def\ddyp{{\mathchoice{\FRAC{1}{d^{\hbox to 2pt{\rm\tiny +\hss}}}{dy}}%
{\FRAC{1}{d^{\hbox to 2pt{\rm\tiny +\hss}}}{dy}}%
{\FRAC{3}{d^{\hbox to 2pt{\rm\tiny +\hss}}}{dy}}%
{\FRAC{3}{d^{\hbox to 2pt{\rm\tiny +\hss}}}{dy}}}}

\def\half{{\mathchoice{\FRAC{1}{1}{2}}%
{\FRAC{1}{1}{2}}%
{\FRAC{3}{1}{2}}%
{\FRAC{3}{1}{2}}}}

\newsavebox{\junk}
\savebox{\junk}[1.6mm]{\hbox{$|\!|\!|$}}

\def\limsup{\mathop{\rm lim{\,}sup}}

\def\argmin{\mathop{\rm arg{\,}min}}

\def\state{{\sf X}}

\def\ustate{{\sf U}}

\def\bfmath#1{{\mathchoice{\mbox{\boldmath$#1$}}%
{\mbox{\boldmath$#1$}}%
{\mbox{\boldmath$\scriptstyle#1$}}%
{\mbox{\boldmath$\scriptscriptstyle#1$}}}}

\def\bfmU{\bfmath{U}}

\def\bfmX{\bfmath{X}}

\def\bfmY{\bfmath{Y}}

\def\bfmhhaY{\bfmath{\hhaY}} 
\def\bfmhhaY{\hbox to 0pt{$\widehat{\bfmY}$\hss}\widehat{\phantom{\raise 1.25pt\hbox{$\bfmY$}}}}

\def\bfmZ{\bfmath{Z}}

\newlength{\dhatheight}

\def\hatheta{{\hat\theta}}

\def\tiltheta{{\tilde \theta}}

\def\lgmath#1{{\mathchoice{\mbox{\large #1}}%
{\mbox{\large #1}}%
{\mbox{\tiny #1}}%
{\mbox{\tiny #1}}}}

\def\Zero{{\mathchoice{\lgmath{\sf 0}}%
{\mbox{\sf 0}}%
{\mbox{\tiny \sf 0}}%
{\mbox{\tiny \sf 0}}}}

\def\clF{\mathcal{F}}   

\def\clZ{\mathcal{Z}}

\newcommand{\eqdef}{\ensuremath{\stackrel{\hbox{\sf\tiny def}}{=}}}

\def\Prob{{\sf P}}

\def\Expect{{\sf E}}

\def\lgmath#1{\mathlarger{#1}}

\def\Zero{{\mathchoice{\lgmath{\sf 0}}%
{\mbox{\sf 0}}%
{\mbox{\tiny \sf 0}}%
{\mbox{\tiny \sf 0}}}}

 \def\epsy{\varepsilon}

\def\formtmp#1#2{{\vskip12pt\noindent\fboxsep=0pt\colorbox{#1}{\vbox{\vskip3pt\hbox to \textwidth{\hskip3pt\vbox{\raggedright\noindent\textbf{#2\vphantom{Qy}}}\hfill}\vspace*{3pt}}}\par\vskip2pt%
\noindent\kern0pt}}

\titleformat\subparagraph[runin]
                 {\normalfont\normalsize\bfseries}
                 {mypar}
                 {0pt}
                 {}{}
\titlespacing\subparagraph{0pt}
                {.1ex minus 0.2ex}
                {.75em}

 \definecolor{pcodecolor}{gray}{0.87}
 \definecolor{shadecolor}{gray}{.95}

\def\zat{0pt}
\newdimen\svparindent
\newenvironment{newshaded}{%
  \MakeFramed
{\FrameRestore}}%
{\endMakeFramed}
\AtBeginDocument{\renewenvironment{svgraybox}%
{\fboxsep=12pt\relax
 \begin{newshaded}\vspace*{7pt}%
 \list{}{\leftmargin=12pt\rightmargin=12pt\topsep=\zat\relax}%
 \expandafter\item\parindent=\svparindent
 \hskip-\listparindent}%
{\gobblepars\vspace*{7pt}\endlist\end{newshaded}}}%

{\endlist\end{newshaded}\pagebreak[3]%
}%

\newtheoremstyle{thm}{12pt}{15pt}%
     {\itshape}
     {}
     {\bfseries}
     {}
     {0pt}
     {\thmname{#1}\thmnumber{ #2.}\thmnote{ \textbf{(#3)}}\quad}

\newcommand\qunderline[1]{\ThisStyle{%
  \ensurestackMath{\stackengine{-0.1pt}{\SavedStyle#1}
  {\SavedStyle\underline{\hphantom{#1}}}{U}{c}{F}{F}{S}}}%
}

\def\uH{\qunderline{H}}

\def\barf{\bar{f}}

\def\psisub#1{\psi_{(#1)}}

\def\csub#1{c_{#1}}

{\end{list}}

\def\ass(#1:#2){(#1\ref{#1:#2})}

\def\ritem#1{
\item[{\sf \ass(\current_model:#1)}]
}

\newenvironment{recall-ass}[1]{%
\begin{description}
\def\current_model{#1}}{
\end{description}
}

\def\sq{\hbox{\rlap{$\sqcap$}$\sqcup$}}
\def\qed{\ifmmode\sq\else{\unskip\nobreak\hfil
\penalty50\hskip1em\null\nobreak\hfil\sq
\parfillskip=0pt\finalhyphendemerits=0\endgraf}\fi}

\newcommand{\blot}{\vrule height 1.1ex width .9ex depth -.1ex }
\def\qedb{\ifmmode\blot\else{\vspace{-.2cm}\unskip\nobreak\hfil
\penalty50\hskip1em\null\nobreak\hfil\blot
\parfillskip=0pt\finalhyphendemerits=0\endgraf}\fi}

\newtheoremstyle{example}{15pt}{20pt}%
     {}
     {}
     {\bfseries}
     {}
     {1pt}
     {\thmname{#1}\thmnumber{ #2.}~\thmnote{\textit{\textbf{#3}}}%
     \\[.15cm]\unskip\nobreak}

\newcounter{rmnum}

\newcounter{anum}

\newlist{EX}{enumerate}{1}
\setlist[EX]{font=\bfseries,label=\thechapter.\arabic*,%
topsep=0pt,%
partopsep=0pt,%
listparindent = 0pt,%
itemsep=3pt,%
parsep=3pt,%
labelwidth=0pt,%
labelsep=8pt,%
itemindent=25pt,%
leftmargin=0pt,%
rightmargin=0pt,%
 }

\newcommand{\field}[1]{\mathbb{#1}}

\def\Re{\field{R}}

\def\transpose{{\intercal}}

\def\argmin{\mathop{\rm arg\, min}}

\def\epsy{\varepsilon}

\def\haY{\widehat{Y}}

\def\hhaY{\hbox to 0pt{$\haY$\hss}\widehat{\phantom{\raise 1.25pt\hbox{Y}}}}

\def\haY{\widehat Y}

 \def\thetaUniquePolicy{\mathcal{C}^\tTheta}
\def\nuEXP{\upnu}

\def\psisub#1{\psi_{(#1)}}

\def\csub#1{c_{#1}}

\def\Qstar{Q^\star}
\def\uQstar{\uQ^\star}

\def\EXP{\mathcal{W}}

\def\tEXP{{\text{\tiny$\EXP$}}}

\def\tEXP{{\text{\tiny$\EXP$}}}

\def\REXP{R^\tEXP}

\def\whamb{\wham{$\bullet$} }

\def\tTheta{{\text{\tiny$\Theta$}}}

\title{Optimistic Training and Convergence of Q-Learning}

\author{Prashant Mehta${}^\dagger$ and Sean~Meyn*
\thanks{${}^\dagger$University of Illinois Urbana-Champaign, Department of Mechanical Science and Engineering (e-mail:  mehtapg@illinois.edu). 
Financial supported in part by the AFOSR award FA9550-23-1-0060 and the NSF award 233613.}%
\thanks{*University of Florida, Department of Electrical and Computer Engineering (e-mail:  meyn@ece.ufl.edu). 
Financial support from ARO award W911NF2010055     and NSF award CCF~2306023
		are gratefully acknowledged.  }%
}

\begin{document}

\thispagestyle{empty}

\maketitle
\thispagestyle{empty}

\def\betag{\beta^\blacklozenge}
\def\thetag{\theta^\blacklozenge}
 \def\cg{c_\blacklozenge}
 \def\Qg{Q^\blacklozenge}
 \def\Ag{A^\blacklozenge}
 \def\feeg{\fee^\blacklozenge}
 \def\upvarpig{\upvarpi^\blacklozenge}
 
\def\sft{\textsf{\footnotesize t}}
\def\thetat{\theta^{\sft}}
 \def\ct{c_{\sft}}
 \def\Qt{Q^{\sft}}
 \def\feet{\fee^{\sft}}
 \def\barft{\barf_{\sft}}

\begin{abstract}

In recent work it is shown that Q-learning with linear function approximation is stable, in the sense of bounded parameter estimates, under the $(\epsy,\kappa)$-tamed Gibbs policy;   $\kappa$ is inverse temperature, and  $\epsy>0$ is introduced for additional exploration.   Under these assumptions it also follows that there   is a solution to the projected Bellman equation (PBE).   Left open is uniqueness of the solution, and criteria for convergence outside of the standard tabular or linear MDP settings.   

The present work extends these results to other variants of Q-learning, and clarifies prior work:   a one dimensional example shows that under an oblivious policy for training there may be no solution to the PBE, or multiple solutions, and in each case   the algorithm is not stable under oblivious training.

The main contribution is that far more structure is required for convergence.    	An example is presented for which the basis is ideal, in the sense that the true Q-function is in the span of the basis.   However,   there are two solutions to the PBE under the greedy policy, and hence also for the $(\epsy,\kappa)$-tamed Gibbs policy for all sufficiently small $\epsy>0$ and $\kappa\ge 1$.

	\wham{Keywords:  reinforcement learning;  optimal control.}
\end{abstract}

	\clearpage
	
	\contentsline {section}{\numberline {1}Introduction}{2}{section.1}%
\contentsline {section}{\numberline {2}Optimism and Stability}{7}{section.2}%
\contentsline {subsection}{\numberline {2.1}Preliminaries}{7}{subsection.2.1}%
\contentsline {subsection}{\numberline {2.2}Challenge with oblivious policies}{8}{subsection.2.2}%
\contentsline {subsection}{\numberline {2.3}Stability with optimism}{9}{subsection.2.3}%
\contentsline {subsection}{\numberline {2.4}Limits of Q learning}{10}{subsection.2.4}%
\contentsline {subsection}{\numberline {2.5}Extensions to SGD}{10}{subsection.2.5}%
\contentsline {section}{\numberline {3}Can We Expect Convergence?}{11}{section.3}%
\contentsline {subsection}{\numberline {3.1}Q-learning in dimension two}{11}{subsection.3.1}%
\contentsline {subsection}{\numberline {3.2}Numerical examples}{12}{subsection.3.2}%
\contentsline {subsection}{\numberline {3.3}Q-learning for MSBE}{13}{subsection.3.3}%
\contentsline {section}{\numberline {4}Conclusions}{14}{section.4}%
	
\section{Introduction}
\label{s:intro}

This paper concerns Q-learning  for a Markov Decision Process (MDP)  with state process $\bfmX$ evolving on a state space $\state$, and input process $\bfmU$ evolving on $\ustate$. In most of the technical results it is assumed that $\state$ and $\ustate $ are finite to avoid discussion of measurability and other technicalities.  

For simplicity only, the technical content is devoted to the discounted-cost optimal control problem:
 For given
discount factor $\disc\in(0,1)$ and cost function $c\colon\state\times\ustate \to \Re$,  the state-action value function  is denoted  
\begin{equation}
\Qstar(x,u) = \min \sum_{k=0}^\infty \disc^k \Expect [ c(Z_k) \mid   Z_0    = (x,u ) ]    
\label{e:Q}
\end{equation}
where   $Z_k = (X_k,U_k)$, and the minimum is over all history dependent input sequences.  
This is the \textit{Q-function} of Q-learning, which solves the Bellman equation, 
\begin{equation}
\Qstar(x,u) = c(x,u) +  \disc \Expect [ \uQstar(Z_1) \mid   Z_0    = (x,u ) ]  
\label{e:BE}
\end{equation}
where $\uH(x) \eqdef\min_u H(x,u)$ for a function $H\colon\state\times\ustate \to\Re$.   

The objective of Q-learning is to obtain an approximate solution among a parameterized class $\{ Q^\theta :  \theta\in\Re^d \}$.
Given any $\theta\in\Re^d$ the $Q^\theta$-greedy policy is defined by
\begin{equation}
\fee^\theta(x)
    = \arg\min_{u} Q^\theta(x,u) \,,  \qquad x\in\state
\label{e:fee_theta}
\end{equation}
with some fixed rule in place in case of ties in the minimum.    In particular, $ \uQ^\theta(x) = Q^\theta(x, \fee^\theta(x)) $.

This paper concerns Q-learning algorithms  motivated by
the following sample path implication of the Bellman equation \eqref{e:BE}:
\begin{equation}
0 =  \Expect [ -\Qstar(Z_k)+ c(Z_k) +  \disc \uQstar(X_{k+1}) \mid   Z_ 0^k]  \,, 
\label{e:BEsp}
\end{equation}
  valid for any admissible input and $  k\ge0  $.  \notes{Need to define admissibility}
The criterion for success of a Q-learning algorithm is  cast   as the   solution to a root finding problem of the form  $\barf(\theta^*) = 0$,  with $\barf\colon\Re^d\to\Re^d$ 
defined as
\begin{equation}
\begin{aligned}
 \barf(\theta) &= \Expect [ f_{k+1}(\theta)]  
\\
  f_{k+1}(\theta) & =
 \elig^\theta_k \{  - Q^\theta(Z_k)  + c(Z_k) + \disc \uQ^\theta(X_{k+1}) \}  
 \end{aligned}
\label{e:barfQ}
\end{equation}  
where 
$\elig^\theta_k$ defines the $d$-dimensional \textit{eligibility vector}, and  the expectation is in steady-state.    

\begin{subequations}

Throughout the paper we assume a linear function class:   $Q^\theta = \theta^\transpose  \psi$  with $\psi\colon\state\times\ustate\to\Re^d$.  
Two standard choices of the eligibility vector in this setting lead to two standard algorithms:
\begin{align}
\elig^\theta_k & = \psisub{k}  \eqdef \psi(Z_k) 
\label{e:PBEelig}
\\
\elig^\theta_k & = \psisub{k} -\disc   \psisub{k+1|k} ^\theta  
\label{e:MSBEelig}
\\
&\quad  \textit{with} \ \ \psisub{k+1|k} ^\theta =    \Expect[ \psi(X_{k+1}, \fee^\theta(X_{k+1}) ) \mid Z_k]  
\nonumber
\end{align}
Observe that the option \eqref{e:MSBEelig} requires a model to obtain the conditional expectation;
a model-free implementation is available   via two time-scale stochastic approximation \cite{avrbordolpat21}.
 \label{e:elig}
\end{subequations}

On solving $\barf(\theta^*) = 0$,  the function
$Q^{\theta^*}$ is known as the solution to 
the \textit{projected Bellman equation}  (PBE)
when using \eqref{e:PBEelig}.  
The alternative \eqref{e:MSBEelig} is motivated by consideration of the  
mean-square Bellman error (MSBE),
\begin{equation}
\Obj(\theta) = \Expect\big[ \{ - Q^{\theta}(Z_k) + c(Z_k) +  \disc \Expect [ \uQ^{\theta}( Z_{k+1}) \mid   Z_ k] \}^2  \big]
\label{e:MSBE_loss}
\end{equation}
and the recognition that $\barf (\theta) = \half \nabla \Obj\, (\theta)$ for any $\theta$ for which $\Obj$ is differentiable.  \notes{whenever the policy is uniquely defined}

\begin{figure*}[h]
  \centering

\includegraphics[width= 1\hsize]{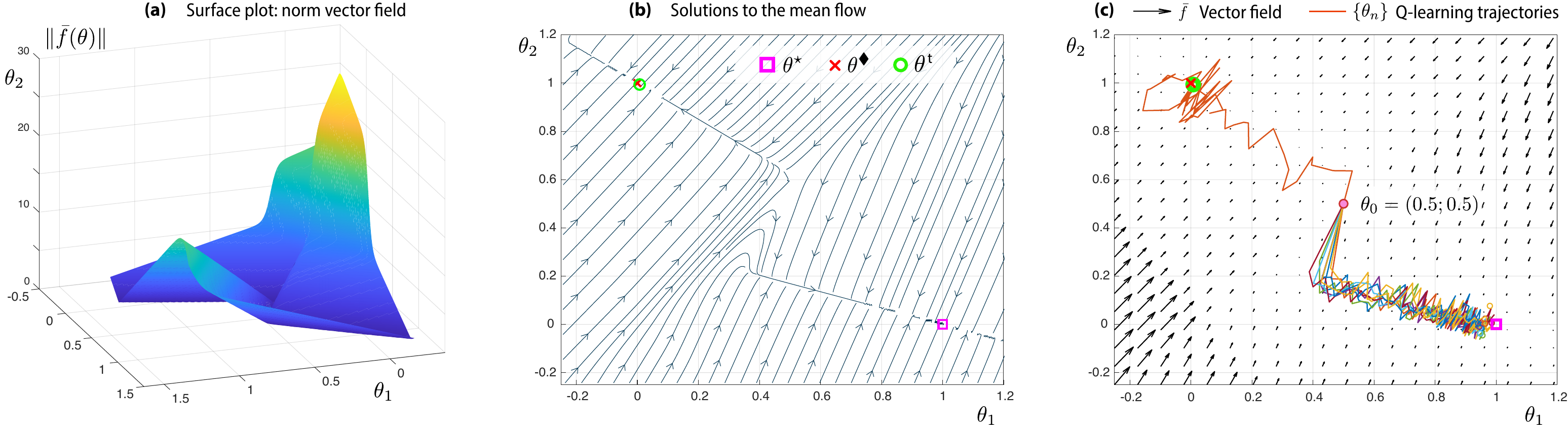}


  \caption{Geometry of the mean flow $\barf$ and sample Q-learning trajectories under the tamed Gibbs policy ($\kappa=20$, $\varepsilon=10^{-3}$).
\textbf{(a)} Norm $\|\barf(\theta) \|$, revealing two basins associated with distinct PBE equilibria.
\textbf{(b)} Mean–flow vectors and solution curves, together with the greedy–policy equilibria ${\theta^\star,\thetag }$ and the tamed–Gibbs equilibrium $\thetat$.
\textbf{(c)} Ten Q-learning trajectories from a common initial condition: one approaches $\thetat$, while the others converge to $\theta^\star$.
}

    \label{f:TwoBE_Solns_Ideal}
\end{figure*}

\begin{subequations}

The   Q-learning algorithms considered in this paper are    generalizations of the original algorithm of Watkins   \cite{watday92a,wat89}:
For initialization $\theta_0\in\Re^d  $,    
\begin{align}
\theta_{n+1} &= \theta_n + \alpha_{n+1}   \Tdiff_{n+1} 
\label{e:Qlambda}
\\
\Tdiff_{n+1} & =  c(Z_n)   + \disc     \uQ^{\theta_n} (X_{n+1})  - Q^{\theta_n}(Z_n) \,  ,
\label{e:BE_theta_n}
\end{align} 
in which  
$\{ \elig_n  \eqdef \elig_n^{\theta_n} \}$  are   the  eligibility vectors
and      $\{\Tdiff_{n+1} \} $ is known as the \textit{temporal difference sequence}.    
The non-negative step-size sequence $\{\alpha_n\}$ is a design choice.   

\label{e:Qgen}
\end{subequations}

When using the eligibility vector \eqref{e:PBEelig} the algorithm is simply called \textit{Q-learning}.
Given the interpretation of $\barf$ as a scaled gradient of the objective 
\eqref{e:MSBE_loss},  when using \eqref{e:MSBEelig} the algorithm is an instance of stochastic gradient descent (SGD), and so we   refer to it as \textit{MSBE-SGD}.

The standard approach to analysis of stochastic approximation algorithms such as \eqref{e:Qgen} is the ODE method, in which solutions to \eqref{e:Qgen} are compared to solutions of the \textit{mean flow},
\begin{equation}
  \ddt \odestate_t = \barf(\odestate_t).
\label{e:ODESA1}
\end{equation}
In particular, the root $\theta^* $ is a stationary point of this ODE.   
Recent results impose exponential asymptotic stability (EAS) of \eqref{e:ODESA1} to obtain convergence of 
general stochastic approximation algorithms in both the almost sure and $L_p$ senses, and also asymptotic statistics such as the 
Central Limit Theorem \cite{borchedevkonmey25}.

The Jacobian of $\barf$   plays a crucial role in this theory:
\begin{equation}
 A(\theta) = \partial_\theta \barf\, (\theta)
\label{e:Atheta}
\end{equation}
In particular we can be assured that the mean flow is \textit{locally} EAS if $A^*\eqdef A(\theta^*)$ is Hurwitz.

For Q-learning, it is known that mean flow \eqref{e:ODESA1} is EAS in the tabular setting of Watkins \cite{tsi94a} and for linear MDPs \cite{yanwan19}.   The lack of theory was part of the motivation for \textit{Zap Q-learning}  
\cite{devmey17a,devmey17b,chedevbusmey20b}, whose mean flow is modified through a matrix gain:    
\begin{equation}
  \ddt \odestate_t = G_t \barf(\odestate_t) \,,  \quad  G_t = - [A(\odestate_t)]^{-1}
\label{e:NRflow}
\end{equation}
resulting in $  \barf(\odestate_t)  = e^{-t}  \barf(\odestate_0) $.  If the matrix $A$ is not everywhere invertible then a pseudo inverse is successful under mild conditions \cite{chedevbusmey20b}.    However, these conditions have not been validated for Q-learning with linear function approximation, even though numerical results have been encouraging.   

\wham{Stability with optimism }

Randomized policies are traditionally used to improve efficiency in training  and efficiency demands that the policy is also parameter dependent, with  \cite{yanwan19} just one example.   We adopt the notation $\feex^{\theta}$ for a randomized policy, which defines the input as follows:  
\begin{equation}
\begin{aligned}
\Prob\{ U_k = u\mid & \clF_k^-;   X_k = x   \}    
=
\feex^{\theta_k} (u \mid x)  
\end{aligned} 
\label{e:RandTraining}
\end{equation}
with $\clF_k^- = \sigma\{  X_i, U_i : i<k  : i\le k    \} $  (a partial history of observations up to iteration $k$).
A training policy is called \textit{oblivious} if $
\feex^{\theta}$ does not depend on $\theta$.

The \textit{tamed Gibbs} policy is defined through the following steps.     It requires specification of 
a pmf  $\nuEXP$ on $\ustate$,  along with scalars
$\epsy,\kappa >0$.   Denote $ \kappa_\theta = \kappa /\sqrt{1 + \| \theta \|^2 } $, 
and then the randomized policy
\begin{equation}
\begin{aligned}
\feex^{\theta}_0 (u \mid x)   =    \frac{1}{ \clZ^\theta_\kappa  (x)}   \exp\bigl( - \kappa_\theta  Q^\theta(x,u)     \bigr)  
\end{aligned}
\label{e:epsyGreedySoftGibbTamed}
\end{equation}
where $ \clZ^\theta_\kappa  (x) $ is a    normalizing constant.
The $(\epsy,\kappa)$-tamed Gibbs policy is then defined by
\begin{equation}
 \feex^{\theta} (u \mid x)  
 =(1-\epsy) \feex^{\theta}_0 (u \mid x)    +  \epsy \nuEXP(u) 
\label{e:tamedGibbs}
\end{equation}
This   approximates the greedy policy $\fee^\theta$ when $\epsy\approx 0$ and $\kappa\approx\infty$.

The main contributions of \cite{mey24,mey23} concern the Q-learning \eqref{e:Qgen} with eligibility vector \eqref{e:PBEelig}.    
With linear function approximation and under an $(\epsy ,\kappa)$-tamed Gibbs policy, the policy parameters can be chosen to ensure the algorithm is stable in the sense of ultimate boundedness:   
 for a deterministic constant $B_{\epsy,\kappa}$ and any initial $\theta_0$,
\begin{equation}
\limsup_{n\to\infty }  \|\theta_n\|  \le B_{\epsy,\kappa}  \quad a.s..  
\label{e:UBQ}
\end{equation}
Under those same assumptions it also follows that there is a solution to the projected Bellman equation (PBE).

%
%
%
%
%

\wham{Contributions.} 
Major issues remain unresolved, notably, (i) whether the solution to the PBE is unique, and (ii) whether convergence can be guaranteed.
This paper advances understanding along these two fronts through both theoretical arguments and constructive counter-examples. Specifically:

\whamb  We provide a simple one-dimensional ($d = 1$) counter-example where a fixed (oblivious) policy causes Q-learning to fail because there is no solution to the PBE, or because the mean flow is not stable. 
This demonstrates that even in the simplest non-trivial linear-approximation setting, convergence cannot be taken for granted. 
This example also serves to illustrate why adopting a tamed Gibbs   training policy ensures stability, but not convergence.

\whamb  Convergence might be expected when using a linear function class  that contains the true Q-function.    
It is shown analytically that this is \textit{not} always true when the dimension of the function class is greater than unity:  the projected Bellman equation may have multiple solutions 
when using an
$(\epsy,\kappa)$-tamed Gibbs  
training policy, subject to the assumptions of  \cite{mey24,mey23} that requires sufficiently small $\epsy$ and large $\kappa$.
This conclusion is   illustrated in   \Cref{f:TwoBE_Solns_Ideal}.

\whamb  The analysis is extended to the MSBE (recall $\Obj(\theta)$ in \eqref{e:MSBEelig}) for which similar conclusions are obtained.  An illustration is provided in   \Cref{f:MSBE2}~(a), showing exactly two local minima for the MSBE \eqref{e:MSBE_loss}.

Overall, this work delivers \textit{explicit evidence of fundamental limitations} in existing convergence/stability claims for linear-approximation Q-learning. In particular, we underscore that oblivious policies can destabilize learning, and that even $\epsy$-greedy policies may fail to guarantee convergence in some settings. These findings  highlight the   need for better guidelines for algorithm design.  

 %
%
%
%
%
%
%

\wham{Literature}

A history of Q-learning with a focus on stability theory may be found in \cite{CSRL}  (see in particular the historical notes sections at the close of chapters 5 and 9),  and further history in \cite{mey24,mey23}.

Note that the tamed Gibbs policy   in \cite{mey24} is defined slightly differently, using 
 \begin{equation}
\kappa_\theta \    \begin{cases}    =  \frac{1}{ \|\theta\|}  \kappa&   \|\theta\|\ge 1
\\				 \ge \half \kappa   &\textit{else}
	\end{cases}  
\label{e:normGibbsBdds}
\end{equation}
chosen so that   the following  holds for all $x,u$:
 \begin{equation}
\text{   $\feex^{r\theta} (u \mid x) =  \feex^\theta(u \mid x) $ for all $r\ge 1$   and $\|\theta\| \ge 1$.}
\label{e:LargeParPolicy}
\end{equation}
The theory from\cite{mey24} carries over since it is sufficient to have convergence:  for the policy 
\eqref{e:epsyGreedySoftGibbTamed}  we have for non-zero $\theta$,
\[
\lim_{r\to\infty } \feex^{r \theta}_0 (u \mid x)   =    \frac{1}{ \clZ^\theta_{\kappa^\infty}  (x)}   \exp\bigl( - \kappa_\theta^\infty  Q^\theta(x,u)     \bigr)  
\]
where $\kappa_\theta^\infty = \kappa/\| \theta\|$.    We opt for \eqref{e:epsyGreedySoftGibbTamed} since we require that $\kappa_\theta$ be analytic in the proof of  
 \Cref{t:ben99}.
 
Conditions for convergence of Q-learning using an oblivious policy are contained in \cite{melmeyrib08}.   
These conditions are very difficult to verify, leading to a number of subsequent works to either relax the assumptions, or modify the Q-learning algorithm.    In 
 \cite{carmelsan20}  it is shown that a two time scale algorithm is convergent subject to conditions on the basis,  but admittedly the algorithm converges to a limit that is not yet fully understood.   Regularization-based approaches guarantee stability by altering the geometry of the Bellman operator \cite{limkimlee22,xigarmom25}, but again the modification in algorithmic objective means more research is needed to understand the performance of the resulting policy.   

The use of truncation has been shown to eliminate divergence \cite{checlamag23}, with convergence  claimed only in the tabular setting.  
  Stability using matrix gain algorithms is the subject of \cite{devmey17b,chedevbusmey20b} where Zap Q-learning was introduced.  However,  since all of these papers assume an oblivious policy results from the present paper imply that convergence cannot be expected without strong assumptions on the basis and the MDP model.  

Geometry of the mean flow vector field summarized in \Cref{t:Aoblivious}  is adapted from   \cite{devmey17a,devmey17b}.  

The existence of a solution to the PBE is a topic of \cite{farroy00} for which existence is established only for a smoothed version of this fixed point equation,
in which the mean flow vector field becomes
\begin{equation}
\begin{aligned}
\barf_{\textsf{s}}(\theta)  &= \Expect [ f_{k+1}(\theta)]  
\\
  f_{k+1}(\theta) & =
 \elig^\theta_k \{  - Q^\theta(Z_k)  + c(Z_k) + \disc Q^\theta(Z_{k+1} ) \}  
 \end{aligned}
\label{e:soothed_barf}
\end{equation}
 where $\{ Z_k = (X_k, U_k) \}$ is the stationary realization of the joint process using the Gibbs policy $\feex^\theta$---see \cite{mey24} for further discussion.   Sufficient conditions for a solution and further history may be found in \cite{limlee25}.

Recall that the algorithm \eqref{e:Qgen}   using the eligibility vector \eqref{e:MSBEelig}  is referred to as MSBE-SGD. 
It has gone by other names in prior research Baird uses \textit{residual-gradient method} \cite{bai95} and more recently it is referred to as 
a \textit{full gradient} method   \cite{avrbordolpat21}.    Neither of these terms are common in today's literature, so we opt for this acronym which we feel more accurately describes the algorithm.

Further history of MSBE-SGD up to 2014 may be found in \cite{danneupet14}.   Convergence of MSBE-SGD is established in  \cite{avrbordolpat21}, but only subject to strong structural assumptions, including realizability and uniqueness of stationary points of the objective \eqref{e:MSBE_loss}. 
  
 Challenges are discussed in the recent work \cite{gotgroshedie21,shasut23}.
 The latter points out numerical ill-conditioning associated with this root finding problem;
 in the language of the present paper, the matrix $A$ appearing in 
 \eqref{e:Atheta} has at least one eigenvalue of magnitude $O( |1-\disc|^2)$.   
 This is addressed in \cite{shasut23} through the introduction of   a Gauss–Newton-inspired algorithm much like  Zap-Zero \cite{CSRL,laubusmey23}.

Finally, note that convergence theory for Q-learning is of limited value if the performance of the resulting policy cannot be assessed. 
In the 1990s, the highly influential work~\cite{sinjakjor95} established convergence of the Q-learning algorithm with basis~\eqref{e:PBEelig} defined via binning. 
The proof relies on stochastic approximation arguments closely paralleling those used in the tabular setting. 
More recent work relaxes several of the restrictive assumptions in this earlier analysis and, more importantly, derives explicit performance bounds that relate the binning architecture to the   performance of the resulting policy~\cite{karsalnuk23,bickaryuk25}.

\wham{Organization}

The remainder of the paper is organized as follows:  
\Cref{s:opt} contains theory surrounding stability of Q-learning, initially focusing on  basis~\eqref{e:PBEelig} (intended to obtain solutions to the projected Bellman equation (PBE)).     The results are extended to MSBE-SGD in 
\Cref{s:SGD_stable}.    Theory surrounding uniqueness of solutions to the PBE are contained in \Cref{s:conv}, along with analogous results for MSBE-SGD.   Conclusions  and directions for future research are contained in \Cref{s:conc}.

\wham{Notation}

 We adopt the shorthand notation,
 \begin{equation}
\begin{aligned} 
& \csub{n} = c(Z_n)\,, \ \  
\psisub{n} = \psi(Z_n)\, ,  \quad n\ge 0
\end{aligned} 
\label{e:subNotation}
\end{equation}
with $\{ Z_n = (X_n; U_n) \}$ the \textit{joint state-input process}.
 
 \whamb
$\fee^\theta$:  $Q^\theta$-greedy policy, with some fixed rule in place in case there is no unique minimizer.

 \whamb  $\thetaUniquePolicy$ is the   region of policy continuity:
  \begin{equation}
  \parbox{.85\hsize}{\raggedright  
  $\theta \in   \thetaUniquePolicy$ if and only if 
  $Q^\theta(x,u)  >  Q^\theta(x,\fee^\theta(x)) $ for all $x$ and all  $ u\neq  \fee^\star(x) $.
  }
\label{e:Sunique}
\end{equation}

 \whamb
$\feex^\theta$:  randomized policy,  always chosen in this paper to smoothly approximate $\fee^\theta$.
This is most frequently the tamed Gibbs policy, written  $\feex^{\theta}_{\kappa,\epsy}$ when the parameters need to be made explicit.

 \whamb
$\uH(x) \eqdef\min_u H(x,u)$ in optimality equations, e.g., \eqref{e:BE}.

 \whamb
Q-learning   notation from \eqref{e:BE_theta_n}:  
step-size $\alpha_n$,
eligibility vector $\elig_n  $,      temporal difference     $\Tdiff_{n+1}$.  

\whamb $\barf \colon  \Re^d\to  \Re^d$, vector field for mean-flow, \eqref{e:barfQ}. 

\whamb  $A(\theta)$ the Jacobian of $\barf$, see \eqref{e:Atheta}, and 
 $A^*\eqdef A(\theta^*)$

\whamb
 $\theta^* \in\Re^d$:  root of $\barf$ when there is only one root.     In \Cref{t:TwoPBE} conditions are presented in which there are two roots.
 
\whamb   $\theta_n$ parameter estimate,   Errors:  $\tiltheta_n = \theta_n-\theta^*$,

 \whamb
$P_u$: controlled transition matrix:
\begin{equation}
\Prob\{ X_{k+1} = x' \mid X_k =x\,,\ U_k =u\} = P_u(x,x')
\label{e:Pu}
\end{equation}

\whamb   A policy $\feex^\theta$ induces a transition matrix 
  \( P_\theta   \)  on $\state$,   and
$T_\theta   $
   the  
transition matrix on $\state\times\ustate$  for the joint state-input process:
\[
T_\theta(z,z') = P_u (x,x')  \feex^\theta(u'\mid x') 
\]
where $z=(x,u)$ and $z'=(x',u')$.

\whamb      \( \uppi^\theta \) is the invariant pmf for  $P_\theta $, 
and $ \upvarpi^\theta $ the invariant pmf for $ T_\theta$ (associated with a policy parameterized by $\theta$).  
Either is  called an invariant probability mass function (pmf) when $\state$ and $\ustate$ are discrete, which is assumed in all of the technical results.

\section{Optimism and Stability}
\label{s:opt}

Until \Cref{s:SGD_stable} the discussion in this section is devoted entirely to the Q-learning algorithm \eqref{e:Qgen} using the  basis~\eqref{e:PBEelig} (intended to obtain solutions to the PBE).   

 The section begins with a description of geometric features of $\barf$ and its Jacobian $A$ for this algorithm.

\subsection{Preliminaries}

The following is used in
\cite{devmey17a,devmey17b}
in a treatment of Zap Q-learning in the tabular setting.   The proof is straightforward.  In particular, the representation \eqref{e:Aoblivious}  may be expressed 
$  A(\theta)
  =
  \Expect_{\upvarpi}\!\bigl[
    \partial_\theta f_{n+1}(\theta)
  \bigr]$   (recall  \eqref{e:barfQ}).  

\begin{lemma}
\label[lemma]{t:Aoblivious}  
Suppose that the input used for training is oblivious,  and  that $\{Z_k = (X_k,U_k) : k\ge 0 \}$ has a unique steady-state pmf  $\upvarpi$ on the finite state-action space  $\state\times\ustate$.   
 Then,  \begin{subequations}

\whamrm{(i)}
The derivative in \eqref{e:Atheta} exists for $\theta\in \thetaUniquePolicy$ and given by
\begin{equation}
A(\theta )= - \Expect_{\upvarpi}  \bigl[  \psisub{n} \{  \psisub{n} - \disc \psisub{n+1}^\theta  \} ^\transpose \bigr] 
\label{e:Aoblivious}
\end{equation} 
with $ \psisub{n+1} ^\theta =    \psi(X_{n+1}, \fee^\theta(X_{n+1}) ) $.    

\whamrm{(ii)}
With $A$ defined in \eqref{e:Aoblivious} for arbitrary $\theta$ we have  
\begin{equation}
\barf(\theta) = A(\theta)\,\theta - b 
\label{e:barfQ_pwlinear}
\end{equation}
with $b = - \Expect_{\upvarpi} \big[\psi(Z_k)\,c(Z_k)\big]$.

\whamrm{(iii)}
Suppose that  the components of $\psi$ are everywhere non-negative. 
Then, $\barf_i$ is a concave function of $\theta$ for each  $i$.  
\end{subequations}
\qed
\end{lemma}

The conclusions of 
\Cref{t:Aoblivious} are very different for a randomized policy that is parameter-dependent.
Just as in the theory of actor-critic algorithms, we must take into account sensitivity of the invariant pmf $\upvarpi_\theta$ with respect to $\theta$.    
Assuming the policy $ \feex^{\theta}$ is continuously differentiable in $\theta$, we denote the \textit{score function} evaluated at the current state action pair by,
\begin{equation}
\Lambda_n(\theta)
\eqdef
\partial_\theta \log \feex^{\theta}(U_n\mid X_n) ,
\end{equation}
 Standard arguments give
$A(\theta)    =
  A_0(\theta) + A_1(\theta) $, 
with 
\[
  A_0(\theta)
=
  \Expect_{\upvarpi_\theta}\!\bigl[
    \partial_\theta f_{n+1}(\theta)
  \bigr] \,,
  \quad
  A_1(\theta)
 =
  \Expect_{\upvarpi_\theta}\!\bigl[
    f_n(\theta)\,
    \Lambda_n(\theta)^\transpose
  \bigr] 
\]
in which the expectations   are
taken in steady state under the policy parameter~$\theta$ (see treatment of actor critic methods in \cite{CSRL}).

\notes{
Might note importance of $A(\theta)$ for asymptotic statistics, and for Zap.
\\
And, my gripe:
Even if the reader is not interested in asymptotic statistics such as the CLT,  
if the MSE converges at a rate slower than $O(1/n)$, such as in Watkins' algorithm with poorly designed step-size, then can we expect useful sample complexity bounds?  
}

\begin{figure}[h]
  \centering
  
  	\includegraphics[width= 0.6\hsize]{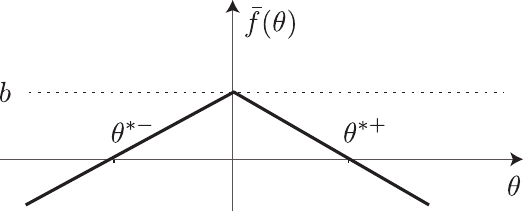}

  \caption{The mean-flow vector field   in case~2 of \Cref{t:Qprobs_Oblivious}.}
\end{figure}

\subsection{Challenge with oblivious policies}

Consider a one-dimensional linear function class, so that
$Q^\theta(x,u)  = \theta\,\psi(x,u)$ for each $x,u$.  
 We also take $x\in\state=\Re$ and $u \in \ustate =  \{0,1\}$. 
Writing $\psi_i (x) = \psi(x,i)$ for each $x , i$ we then have  $Q^\theta(x,u)
    = \theta \big[\,  (1-u) \,\psi_0(x) + u\,\psi_1(x) \big]$, and the $Q^\theta$-greedy policy will be expressed,
\begin{equation}
\fee^\theta(x) 
=
\begin{cases}  \fee^\oplus(x) 
=
\ind\{\, \psi_0(x) \ge \psi_1(x) \,\}, & \theta \ge 0, \\[6pt]
 \fee^\ominus(x) 
=
\ind \{\, \psi_0(x) \le \psi_1(x) \,\}, & \theta < 0.
\end{cases}
\label{e:1dfee-theta}
\end{equation}
Note that arbitrary tie-breaking choices are inherent in this representation, particularly when $\theta=0$.

It is assumed that $\psi(x,u)\ge 0 $  for all $x,u$,   and also bounded and continuous in $x$.

Consider the following implications of \Cref{t:Aoblivious}, subject to an oblivious training policy
 for which the the joint input-state process is ergodic.     
Denote $R \eqdef \Expect_{\upvarpi} \big[\psi(Z_k)^2\big]$,  and 
\[
\begin{aligned}
 R_{1,\oplus} &= \Expect_{\upvarpi} \big[\psi(Z_k)\,\psi(X_{k+1} ,    \fee^\oplus( X_{k+1} ) \big]
 \\
 R_{1,\ominus} &= \Expect_{\upvarpi} \big[\psi(Z_k)\,\psi(X_{k+1} ,    \fee^\ominus ( X_{k+1} )    )\big]
\end{aligned}
\]

The proof of the proposition that follows rests on the representation \eqref{e:Aoblivious} in this special case:  
\begin{equation}
A(\theta)
    = -R +
    \begin{cases}
        \disc\,  R_{1,\oplus}  & \theta\ge0,\\[4pt]
           \disc\,  R_{1,\ominus} & \theta<0 .
    \end{cases}
\label{e:AsimpleOblivious}
\end{equation}
Concavity of \(\barf\) combined with \eqref{e:barfQ_pwlinear}
 implies that  
$A(\theta)\le A(-\theta) $  for $  \theta>0$.

 \Cref{t:Aoblivious} gives $ \barf(\theta) = A(\theta)\,\theta + \Expect_{\upvarpi} [  \csub{k}  
\psisub{k}    ] $.   Each part of the proof of  
 \Cref{t:Qprobs_Oblivious} then follows from \eqref{e:AsimpleOblivious}. 

\begin{figure*}
  \centering
  
  	\includegraphics[width= \hsize]{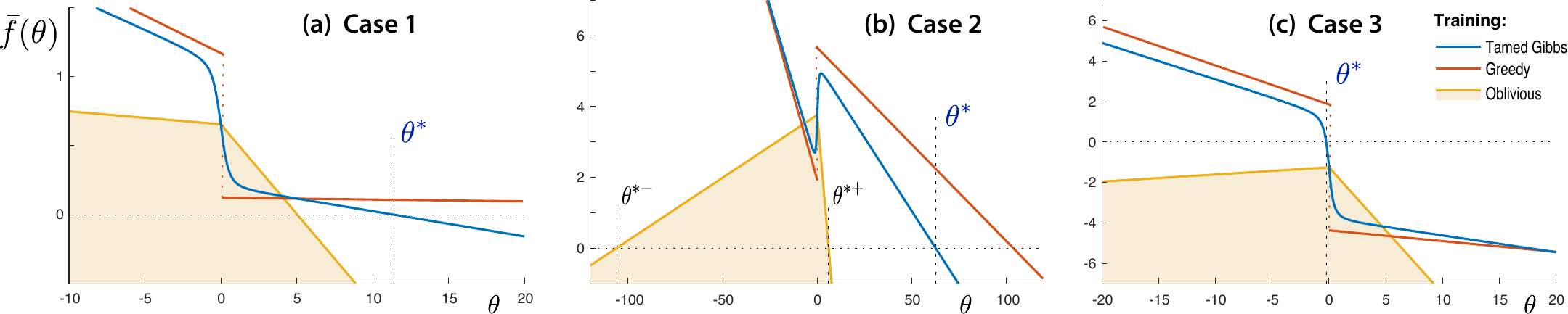}  

  \caption{The   function \(\barf\) illustrating each of the three  cases of \Cref{t:Qprobs_Oblivious}, based on \eqref{e:linEx}, with parameters
    \textbf{(a)}  $m=-10$,  $r=0.1$,  $w = 1$.
    \textbf{(b)}  $r=-m=1$,  $w = 10$.
    \textbf{(c)}  $r=-m=1$,  $w = -10$.
    In each case the mean flow is EAS under the tamed Gibbs policy,  but the oblivious policy is EAS only in case 1.   
}
    \label{f:case123}

\end{figure*}

 \begin{proposition}
 \label[proposition]{t:Qprobs_Oblivious}
 Suppose that the input used for training is oblivious,  and  that $\{Z_k = (X_k,U_k) : k\ge 0 \}$ has a unique steady-state distribution $\upvarpi$ on $\state\times\ustate$.      
  Then, the following are possible outcomes for this Q-learning architecture:
 \whamrm{Case 1:} If \(\disc R_{1,\ominus} < R\), then there is a unique solution to the PBE
      that is EAS for the mean flow.

\whamrm{Case 2:}  If \(\disc R_{1,\ominus} > R\) and
      \(\Expect_{\upvarpi} [\psi(Z_k)c(Z_k)] < 0\), then there are two solutions
      to the PBE, denoted \(\theta^{*-}<0\) and \(\theta^{*+}>0\).
      Solutions to the mean flow diverge to \(-\infty\) if
      \(\odestate_0 < \theta^{*-}\).

\whamrm{Case 3:}  If \(\disc R_{1,\ominus} > R\) and
      \(\Expect_{\upvarpi} [\psi(Z_k)c(Z_k)] > 0\), then there is no solution
      to the PBE.
 \qed
\end{proposition}

An example illustrating the conclusions of the proposition is provided next,  and also  demonstrates that cases~2 and~3 can be ruled out
when using a suitably ``optimistic policy''.
 The state process $\{  X_k : k\ge 0\}$ is Gaussian, evolving as
\begin{equation}
X_{k+1} = \alpha X_k  +   W_{k+1}  
\label{e:linEx}
\end{equation} 
with $\{W_k \}$ i.i.d.\ 
 $N(0,1)$, and independent of the i.i.d.\  input $\{  U_k : k\ge 0\}$.   
 The oblivious policy uses $\Prob\{ U_k=1 \} = 1- \Prob\{ U_k=0\} =1/2$,

   The cost is $c(x,u) = x^2 + w u^2$,  with $w\in\Re$  (possibly negative).

With   $Q^\theta(x,u)  =  \theta \psi(x,u) $ we write $\psi_i (x) = \psi(x,i)$ for each $x , i$,  choosing 
$\psi_1(x) = e^x/(1+e^x) $ and   $ \psi_0(x) =   r + \psi_1(m - x)$  for scalar parameters $r, m$.

    
 \Cref{f:case123} shows the resulting mean-flow vector field for three choices of parameters $r,m,w$,
 and with  $\alpha = 0.5$ and $\disc=0.9$.   It is evident from the three plots labeled ``oblivious'' that model parameters can be chosen to obtain each of the three cases of  \Cref{t:Qprobs_Oblivious}.

\subsection{Stability with optimism}

 Also shown in  \Cref{f:case123} are the mean-flow vector fields for the $Q^\theta$-greedy policy, and the  tamed Gibbs policy using $\kappa =5$ and $\epsy=0$.   The function $\barf$ is continuous and strictly decreasing in cases (a) and (c) for the tamed-Gibbs policy,  which implies that $\theta^*$ is EAS.   
 In all three cases there is $b_0>0$, $\epsy_0>0$ such that under the tamed Gibbs policy,
\begin{equation}
\theta^\transpose \barf(\theta)  \le -\epsy_0 \| \theta \|^2 \ \ \textit{when} \ \   \|\theta \|\ge b_0
\label{e:QuadLyapQ}
\end{equation} 
 This implies ultimate boundedness of the mean flow, and   further steps lead to a proof of  ultimate boundedness of the parameter estimates in the form 
 \eqref{e:UBQ}.
   This is an illustration of \Cref{t:Qtamed} that follows.

When $\epsy=1$ in  \eqref{e:tamedGibbs} we write $U_k = \EXP_k$ for all $k$,  where  $\{ \EXP_k \}$ is an i.i.d.\ sequence on $\ustate$ with marginal $ \nuEXP$.   Let    $\upvarpi_\tEXP$ denote the resulting steady state pmf for $\{Z_k\}$.
    
 \begin{proposition}
 \label[proposition]{t:Qtamed}
 Consider the Q-learning recursion \Cref{e:Qgen} based on the PBE eligibility vector \eqref{e:PBEelig}.     Suppose that the oblivious policy obtained using \eqref{e:tamedGibbs}  with $\epsy = 1$ gives rise to  an aperiodic and uni-chain Markov chain $\{Z_k\}$,  with unique invariant pmf $\upvarpi_\tEXP$, 
and    the autocorrelation matrix  below is positive definite:
\begin{equation}
 \REXP = \Expect_{\upvarpi_\tEXP}  \bigl[  \psi(X_{n}, \EXP_n)    \psi(X_{n}, \EXP_n )^\transpose\bigr]  
\label{e:AutocorrEXP}
\end{equation}
Then, there is $\epsy_\disc >0$ such that for any $\epsy \in (0, \epsy_\disc)$ there is $\kappa_{\epsy,\disc}>0$ for which
 the following hold using the  $(\epsy,\kappa)$-tamed Gibbs policy with $\kappa\ge  \kappa_{\epsy,\disc}$:   

\whamrm{(i) \it  Stability:}    The inequality   \eqref{e:QuadLyapQ} holds for positive constants  $\epsy_0 $, $ b_0$,
  depending on $\kappa  ,\epsy$,
 and there is $ B_{\epsy,\kappa} <\infty$ such that \eqref{e:UBQ} holds.

\whamrm{(ii) \it Solution to the PBE:}    There exists a solution $\theta^*$ to  $  \barf(\theta^*) = \Zero$,   with $\barf$ the mean flow \eqref{e:barfQ}.
 \end{proposition}

\wham{Proof:}
The conclusions follow from \cite[Theorem IV.1]{mey24}, whose proof is based on establishing 
\eqref{e:QuadLyapQ};  the bound follows from 
    the construction of $\epsy_1>0$ and $b_1<\infty$ such that 
\begin{equation}
y^\transpose  A (\theta ) y    \le  - \epsy_1 \| y\|^2 \,, \quad y\in\Re^d\, ,  \  \| \theta \|\ge b_1\, .
\label{e:barfCoercive}
\end{equation}
 \qed

\subsection{Limits of Q learning}

%
%

We can establish that Q-learning has convergent sample paths only  for dimension 1:

\begin{proposition}
\label[proposition]{t:ben99}
Consider the Q-learning algorithm subject to the assumptions of \Cref{t:Qtamed} with $d=1$.  If 
$\epsy \in (0, \epsy_\disc)$  and $\kappa\ge  \kappa_{\epsy,\disc}$,  then the mean flow vector field $\barf$ associated with the 
$(\epsy,\kappa)$-tamed Gibbs policy   has a finite number of roots $\Uptheta^* = \{ \theta^{*1},\dots, \theta^{*m}\}$  where the integer $m\ge 1$ and the values of the $m$ parameters may depend upon   $\disc, \epsy, \kappa  $.   

Moreover, with probability one,   the sequence $\{\theta_n : n\ge 0\}$  converges to a (possibly sample path dependent) element of the finite set $\Uptheta^*$.
\end{proposition}
\notes{I haven't yet ruled out an interval of roots!}

\wham{Proof}  
Under the assumptions of \Cref{t:Qtamed} the function  $\| \barf \|^2$ is coercive due to 
\eqref{e:barfCoercive}, and it is also \textit{semianalytic} in the sense that $
\{x\in\Re : h(x)\ge c \} $  is semianalytic in $\Re$  \cite{biemil88}.     
These conclusions combined imply that $\barf$ has a finite number of roots. 

To establish convergence we apply 
\cite[Theorem 2, Ch.~2]{bor20a}, based on
\cite{ben99a}, which tells us that $\{\theta_n \}$ converges almost surely to one of the \textit{connected internally chain transitive invariant sets} of the mean flow   \eqref{e:ODESA1}.  \qed
 
 Without further assumptions, in
  dimension $d=2$ the 
 Poincar\'e-Bendixson theorem combined with \cite{ben99a}
  tells us that 
  $\{\theta_n \}$  may converge to a periodic orbit,  and more complex behaviors are possible for $d>2$.

\subsection{Extensions to SGD}
\label{s:SGD_stable}

Recall the MSBE $\Obj$ was defined in \eqref{e:MSBE_loss}.
 \begin{proposition}
 \label[proposition]{t:Q-SGDtamed}
 Consider the Q-learning recursion \Cref{e:Qgen} based on the MSBE eligibility vector \eqref{e:MSBEelig}.   
 Assume the same assumptions as in \Cref{t:Qtamed} for the tamed Gibbs policy.  In particular, that  $\REXP >0$ with autocorrelation matrix defined in 
 \eqref{e:AutocorrEXP}.   Then, there is $\epsy_\disc >0$ such that for any $\epsy \in (0, \epsy_\disc)$ there is $\kappa_{\epsy,\disc}>0$ for which
 the following hold using the  $(\epsy,\kappa)$-tamed Gibbs policy with $\kappa\ge  \kappa_{\epsy,\disc}$:   

\whamrm{(i) \it  Stability:}    The inequality   \eqref{e:QuadLyapQ} holds for positive constants  $\epsy_0 $, $ b_0$,
  depending on $\kappa  ,\epsy$,
 and there is $ B_{\epsy,\kappa} <\infty$ such that \eqref{e:UBQ} holds.

\whamrm{(ii) \it Stationary points for the MSBE:}    There exists a solution $\theta^*$ to  $   \nabla \Obj\, (\theta^*) = 0$.
\end{proposition}

\wham{Proof}
 The mean flow vector field has a form similar to  \eqref{e:barfQ_pwlinear}:
$ \barf(\theta) = A(\theta)\,\theta - b(\theta)  $ in which $A = A_{\textsf{\tiny BE}} -\disc^2 K^\theta$ with
\[
 \begin{aligned}
 A_{\textsf{\tiny BE}}(\theta ) &= - \Expect_{\upvarpi_\theta}  \bigl[  \psisub{n} \{  \psisub{n} - \disc \psisub{n+1}^\theta  \} ^\transpose \bigr] 
 \\
K^\theta &=\Expect_{\upvarpi_\theta}  \bigl[ \psisub{n+1}^\theta \{  \psisub{n+1}^\theta  \} ^\transpose \bigr] 
  \end{aligned}
\]
And $b = - \Expect_{\upvarpi_\theta} \big[\psi(Z_k)\,c(Z_k)\big]$ is a bounded function of $\theta$.   

This representation implies that \eqref{e:barfCoercive} holds with this new vector field:  
$ y^\transpose  A (\theta ) y    \le  - \epsy_1 \| y\|^2  $ for all  $ y\in\Re^d$   and    $\| \theta \|\ge b_1$,   which implies ultimate boundedness of the mean flow.   

We omit the proof of (ii) since it is identical to the proof of existence in \cite{mey24}. 
\qed

\section{Can We   Expect Convergence?}
\label{s:conv}

On reviewing \Cref{f:case123}, observe that $\barf$ is monotone decreasing for both tamed-Gibbs and oblivious policies in Case~1.   This case is special because both the cost function and the basis take on non-negative values.   One might hope that this desirable conclusion   extends to dimensions beyond one.   Here we examine the most ideal situation with two-dimensional parameterization, in which the true Q-function lies within the span of the basis.  We prove analytically that convergence cannot be guaranteed in general when using either the tamed Gibbs policy or an $\epsy$-greedy policy.

We begin with theory for the projected Bellman error, which is the context for \Cref{f:case123}.   Similar results are then obtained for MSBE-SGD.

\subsection{Q-learning  in dimension two}

The construction of a two-dimensional basis begins by introducing a second cost function $\cg$.   With $\Qstar$ defined in \eqref{e:Q} and $\Qg$ the Q-function  obtained using $\cg$,  we define a two dimensional basis,
\begin{equation}
\psi(x,u) = (\Qstar(x,u);  \Qg(x,u) )
\label{e:PBEtwoBasis}
\end{equation}
 It is evident that $\theta^\star \eqdef (1;0)$  results in a solution to the Bellman equation,   $Q^{\theta^\star} = \Qstar$.

Let $\barf_{\kappa,\epsy}$ denote the vector field \eqref{e:barfQ}  when the input for training is chosen using the $(\epsy,\kappa)$-tamed Gibbs policy, and  
  $\barf_0$   the vector field \eqref{e:barfQ}  when the input is chosen using the $Q^\theta$-greedy policy.    Assumption (a) in the following is imposed to ensure that \eqref{e:barfQ}  is well defined in every case.

\begin{proposition}
\label[proposition]{t:TwoPBE}

Consider an MDP with finite state and action space.   
It is assumed that the Q-learning algorithm with basis   \eqref{e:PBEtwoBasis} satisfies the assumptions of \Cref{t:Qtamed}, and the following additional assumptions

   \whamrm{(a)}  For any $\theta$, 
 each of the policies $\{ \fee^\theta , \feex^\theta :  \theta\in\Re^2 \}$    induces a Markov chain  $\bfmZ$ with unique invariant pmf.  
   \whamrm{(b)}  The
  pair of autocorrelation matrices are full rank:
\[
R_\star = \Expect_{\upvarpi_\star}[  \psisub{n} \psisub{n} ^\transpose]  
\quad
R_\blacklozenge = \Expect_{\upvarpig}[  \psisub{n} \psisub{n} ^\transpose]  
\]
where $\upvarpi_\star$ is the invariant pmf on $\state\times\ustate$ obtained using the optimal policy, and $\upvarpig$ is the invariant pmf obtained using the $\Qg$ greedy policy: $\fee^\blacklozenge(x) =\argmin_u \Qg(x,u)$.

\whamrm{(c)}  
There is $\betag>0$ such that $\barf_0(\thetag ) =0$ with $\thetag  =(0;\betag)$.

   \whamrm{(d)}
 The two greedy policies associated with $\theta^* = (1;0)$ and $\thetag $ 
    are uniquely defined:  $\theta^*,\thetag \in   \thetaUniquePolicy $.


Then,  the constant  $\epsy_\disc >0$ can be reduced, and   $\kappa_{\epsy,\disc}>0$ increased such that the following hold 
for some fixed constant $b_0>0$ 
when using the
 $(\epsy,\kappa)$-tamed Gibbs policy, provided 
 $\epsy \in (0, \epsy_\disc)$ and $\kappa\ge  \kappa_{\epsy,\disc}$:

 \whamrm{(i)}   
 There exists $\thetat\in\Re^2$ solving 
 $\| \thetat - \thetag \| \le b_0 [ \epsy + 1/\kappa]$
 and 
 \begin{equation}
0 = \barf_{\kappa,\epsy} (\thetat)  =  \Expect_{\upvarpig} [   \psisub{k} \{   c(Z_k)  - \betag \, \cg  (Z_k) \} ]
\label{e:SecondPBEsoln}
\end{equation}

 \whamrm{(ii)}   The two Jacobians $A^\star =  \partial_\theta \barf \, (\theta^\star ) $  and 
 $A^\blacklozenge = \partial_\theta \barf \, (\thetat) $  are Hurwitz, so that both  $\theta^\star$ and $\thetat$ are locally asymptotically stable equilibria for the mean flow.  
 \qed
\end{proposition}

Under (c)-(d) we have $\theta^*,\thetag\in \thetaUniquePolicy$  (recall \eqref{e:Sunique}).   This is crucial in the proof of the proposition, and also permit an extension of the conclusions  to the $\epsy$-greedy policy (corresponding to $\kappa = \infty$).

 \wham{Proof of \Cref{t:TwoPBE}:}    
 First observe that by the  dynamic programming equation that $\Qg$ satisfies we have for any (possibly parameter dependent)  training policy,
 \[
 \barf ((0;\beta) )  =  \Expect [   \psisub{k} \{   c(Z_k)  - \beta \, \cg  (Z_k) \} ] 
 \]
 where the expectation is in steady-state.   This explains the second inequality in \eqref{e:SecondPBEsoln}.
 
 Under the uniqueness assumption  (d),  there exists $\epsy>0$ such that the following hold under the $Q^\theta$-greedy policy:
 \[
 \barf_0(\theta) = \begin{cases}   
 					\Ag_0 (\theta - \thetag)  &   \|\theta - \thetag\| \le \epsy
					\\ 
				A^*_0 (\theta - \theta^*)  &   \|\theta - \theta^*\| \le \epsy
 			\end{cases} 
 \]
 with $\Ag_0 = \partial\barf_0\, (\thetag)$ and $A^*_0 =  \partial\barf_0\, (\theta^*)$.
 
 Next, under the rank condition (b) each of the matrices $\Ag_0 ,A^*_0 $ are Hurwitz since they correspond to the Jacobians for on-policy TD-learning---see \cite[Proposition 9.8]{CSRL} which is based on  \cite{tsivan97}.  
  
The proof of (i) follows from the Implicit Function Theorem, which is justified because the two Jacobians are full rank.     
The Hurwitz property is preserved under small perturbations, which implies (ii).   
\qed

\subsection{Numerical examples}
\label{s:numPBE}

The proposition does not directly provide an example that $\barf$ has multiple roots.   The next step is to construct cost functions and a controlled transition matrix for which (a)--(c) are satisfied.

Code was constructed to produce a controlled transition matrix and non-negative cost functions $c$ and $\cg$ on $\state\times\ustate$ for which the assumptions of \Cref{t:TwoPBE} hold.   In the results surveyed here, the code used $\state = \ustate = \{0,1\}$ and discount factor   $\disc  = 0.75$.   We are then assured of a solution to 
\eqref{e:SecondPBEsoln};  on scaling $\cg$ we obtain without loss of generality $\betag=1$.   
In one set of experiments the following were obtained:  
\[
P_0 =
 \begin{bmatrix}
    0.2164   & 0.7836\\
    0.8437   & 0.1563
\end{bmatrix}
\quad
P_1 =
 \begin{bmatrix}
    0.6077  &  0.3923 \\
    0.3130   & 0.6870
\end{bmatrix}
\]
And, regarding a cost function as a $2\times 2$ matrix in which $c(x,u)$ denotes the $(x+1, u+1)$ entry,  and similar notation for Q-functions,
\[
\begin{aligned}
c  &=
 \begin{bmatrix}
    3.3346   & 1.0275 \\
    2.3374 &   0.2424
\end{bmatrix}
\quad
&&
\cg &=
 \begin{bmatrix}
    3.3346  &  4.4942 \\
    0.8280   & 0.2424
\end{bmatrix}
\\ 
Q^*  & =
 \begin{bmatrix}
    4.9351  &  2.9238  \\
    4.4121   & 1.9160
\end{bmatrix}
\quad
&&
\Qg  &=
 \begin{bmatrix}
    3.5603  &  8.5661 \\
    5.4102   & 3.6770
\end{bmatrix}
\end{aligned}
\]
This resulted in 
$\fee^\blacklozenge(x) = x$ and $\fee^\star(x) \equiv 1$.

\begin{wrapfigure}[22]{r}{0.50\hsize}
	\vspace{-1.0em}					 
	\centering     
	\includegraphics[width=\hsize]{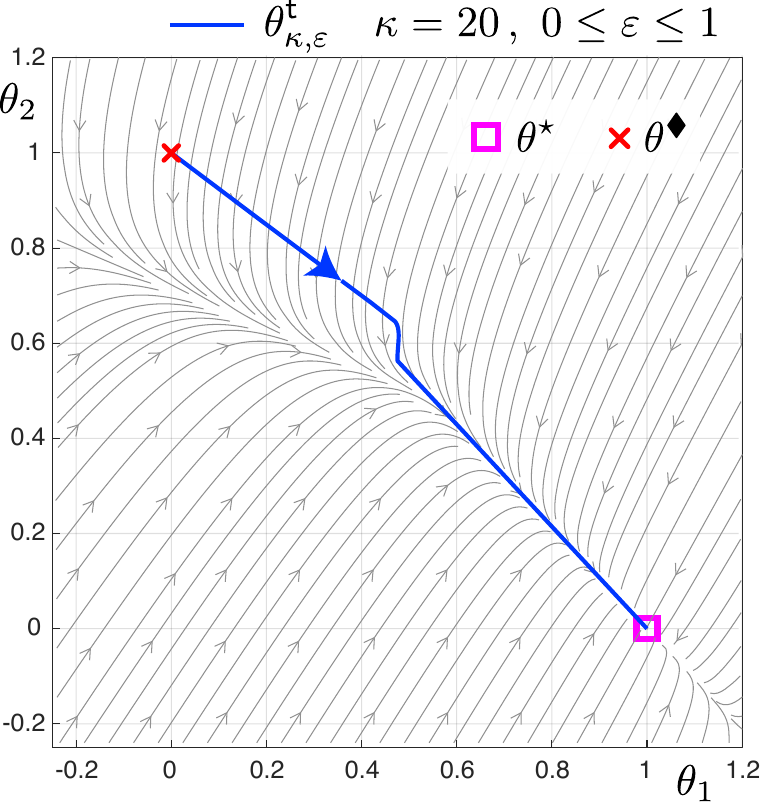}
	\vspace*{-5.5mm} 
  \caption{Path of $\thetat$ as a function of $\epsy$ with $\kappa = 20$:   a continuous path satisfying  
   $\thetat_{\kappa,\epsy} \to \theta^\star$ as $\epsy\uparrow 1$.  }
   
  \label{f:PBE_epsy}
 \end{wrapfigure}

For large $\kappa$ and small $\epsy$ there are exactly two solutions to the PBE, each close to the values $\theta^* = (1;0)$ and $\thetag = (0;1)$.  In particular,  for   $\kappa   = 20$ and $\epsy = 10^{-3}$ we find that
 $\thetat= (     0.0076 ;     0.9942 )$ solves the PBE, which  is indeed close to $\thetag$.  

\Cref{f:PBE_epsy} shows a plot of $\thetat$ as a function of $\epsy\in [0,1]$ with   $\kappa   = 20$ (fixed).   Note that the oblivious policy is best for this example:   $\thetat $ is unique when $\epsy=1$ and approximates $ \theta^*$.   Moreover, the grey curves indicate the solution to the mean flow when  $\epsy=1$, showing that $\theta^*$ is globally asymptotically stable.  It is in fact EAS since the matrix $\partial \barf \, (\theta^*)$ is Hurwitz.

\wham{Best of $M$}   There is no theory available to explain the success of the oblivious policy in this example, and there is ample theory that oblivious policies should be avoided in most cases---traditionally, parameter-dependent policies are celebrated for potential efficiency gains, and we now know that they can be designed to ensure algorithmic stability.

Given that we are not assured of convergence, it is sensible to  perform multiple runs, estimating performance of the resulting policies, and then choosing the best parameter observed in the totality of experiments.

 \begin{figure}[h]
  \centering
  
  	\includegraphics[width= 0.75\hsize]{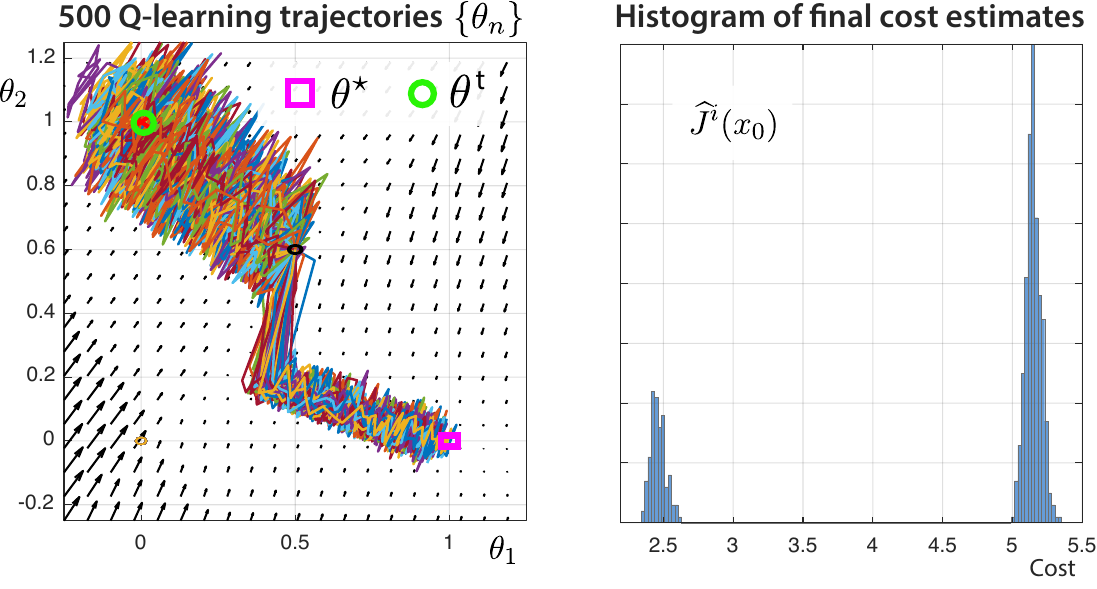}
   
  \caption{Shown on the left hand side are the trajectories of Q-learning from 500 independent experiments to obtain estimates $\{\hatheta^i : 1\le i\le M\}$.     The histogram on the right hand side shows estimates of the 
   discounted cost  from the $Q^{\hatheta^i}$-greedy policy for each parameter.
   }
  \label{f:500Q}
\end{figure}

An illustration is provided in  \Cref{f:500Q}, showing experiments using  $\kappa   = 20$ and $\epsy = 10^{-3}$, for which there are exactly two solutions to the PBE.    The figure shows $M=500$ independent runs for   Q-learning with common initial condition  $\theta_0  = (0.5; 0.6)$,  chosen so that the limit points were roughly equally divided between $\theta^\star$ and $\thetag$.    Each Q-learning trial used a short runlength of $N=5\times 10^3$.   The $M$ runs result in parameter estimates $\{\hatheta^i : 1\le i\le M\}$.   For each estimate,  the discounted cost was estimated  using Monte-Carlo, $J(x_0) = 
 \sum_{k=0}^\infty \disc^k \Expect [ c(Z_k) \mid   X_0    =  x_0 ]    $ with $x_0 = (0;0)$.  A histogram of the results is shown on the right hand side of  \Cref{f:500Q}.    \notes{See Dec 11 version for notes on estimating $J$}

\begin{figure*} 
  \centering
  
  	\includegraphics[width= 0.9\hsize]{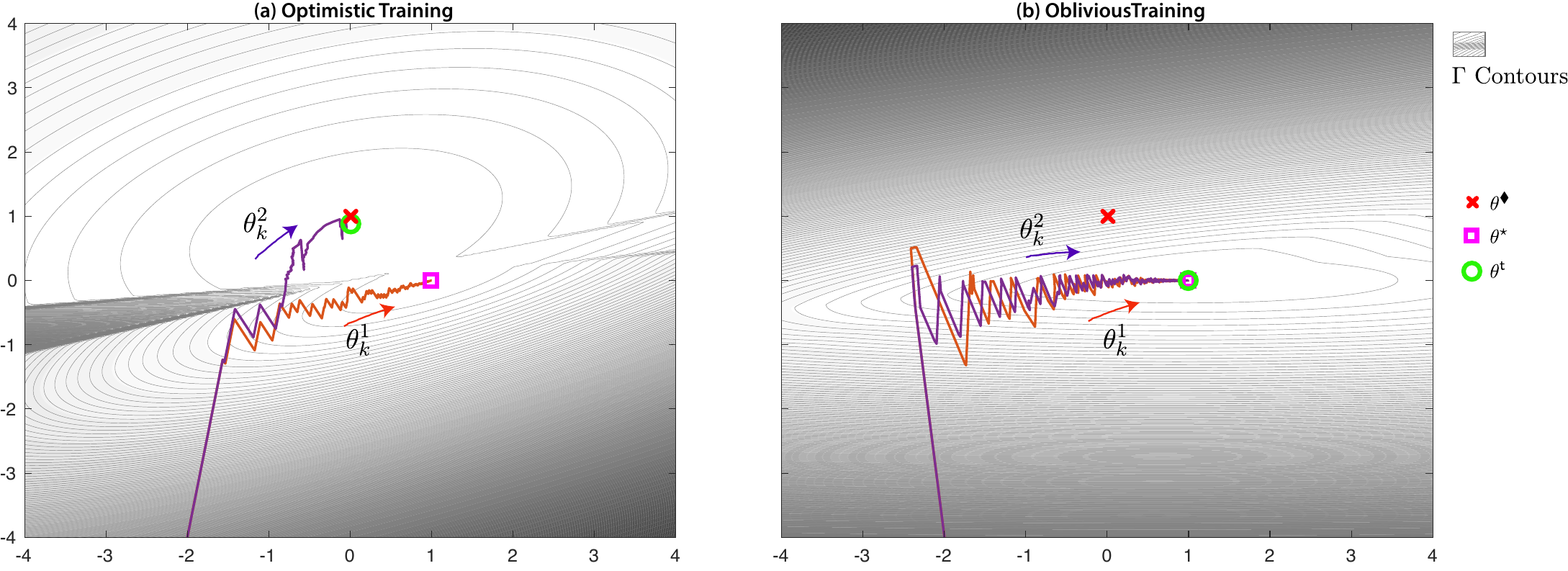}  
   
  \caption{Two independent   runs using  MSBE-SGD  from a single initial conditions using the tamed Gibbs policy.
  (a)  tamed Gibbs policy using $\kappa=20$, $\varepsilon = 10^{-3}$.   (b)  Oblivious policy using $\varepsilon = 1$.}
  \label{f:MSBE2}
\end{figure*}

\subsection{Q-learning for MSBE}

We turn next to MSBE-SGD, defined as \eqref{e:Qgen}  using the eligibility vector \eqref{e:MSBEelig}.   \Cref{t:TwoPBE} has an exact analog.  Rather than use up space with a formal statement, we illustrate the conclusions with numerical examples.

Each of the two plots shown in   \Cref{f:MSBE2} are based on the same controlled Markov model and cost functions used in     \Cref{f:PBE_epsy,f:500Q}.    Shown in (a) are two selected trajectories from Q-learning using the $(\epsy,\kappa)$-tamed Gibbs policy
with and   $\kappa   = 20$ and $\epsy = 10^{-3}$.    We see that there are two solutions to $\barf(\theta) =0$, which are stationary points for the   MSBE  objective $\Obj(\theta)$.   The Q-learning algorithm is convergent to one of these two values with probability one.

 \Cref{f:MSBE2}~(b) shows results   for the oblivious policy obtained with $\epsy=1$.      
   As in the example considered in \Cref{s:numPBE}
we find that $\theta^* = (1;0)$ is the unique root of $\barf$ in this case, and the Q-learning algorithm is convergent to this ideal value from each initial condition.

\section{Conclusions}
\label{s:conc}

The science of reinforcement learning is lagging far behind the success seen in practice.   This paper underscores the wealth of open problems in the theory of Q-learning.   While actor-critic methods have a far stronger theory,  those algorithms promoted today  are designed to reduce the high variance seen in the original formulations, but in doing so lose the original motivation that they represent unbiased stochastic approximation for stochastic gradient descent  \cite[Ch.~10]{CSRL}.

Much of our current research focuses on alternatives to Q-learning such as LP methods 
\cite{mehmey09a,lumehmeyneu22,lumey23a} 
or techniques inspired by mean field games  
\cite{jostagmehmey22,jostagmehmey22,joschatagmehmey25}.

\bibliographystyle{abbrv}
\def\cprime{$'$}\def\cprime{$'$}

%
%
%
%
	
  \end{document}